\RequirePackage{fix-cm}
\documentclass[twocolumn]{svjour3}          
\smartqed  

\usepackage[T1]{fontenc}
\usepackage{lmodern}
\usepackage{cite}
\usepackage{graphicx}
\usepackage{ragged2e}
\usepackage[tight,footnotesize]{subfigure}
\usepackage{rotating}
\usepackage{graphicx}
\usepackage{amsmath,amssymb} 
\usepackage{array}
\usepackage{multirow}
\usepackage{colortbl}
\usepackage{amsfonts}
\usepackage{pifont}
\usepackage{xspace}
\usepackage{etoolbox}
\usepackage{overpic}
\usepackage{color}
\usepackage{microtype}

\usepackage{amsmath, amssymb} 
\usepackage{xcolor}           
\usepackage[ruled,vlined]{algorithm2e} 

\usepackage{booktabs}
\usepackage{bm}

\newcommand{\orcid}[1]{\href{https://orcid.org/#1}{\includegraphics[width=10pt]{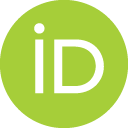}}}

\usepackage{hyperref}
\hypersetup{breaklinks=true,citecolor=blue, colorlinks}

\graphicspath{{./Imgs/}}
\DeclareGraphicsExtensions{.pdf,.jpg,.png}

\usepackage{silence}
\journalname{Research Article}

\begin{document}
\providecommand{\bibcommenthead}{}

\title{HDVS: semi-supervised semantic segmentation via heterogeneous dual-branch voting supervision}

\titlerunning{HDVS: heterogeneous dual-branch voting supervision}        

\author{Yongqi Shan \orcid{0009-0004-3469-5231}        \and
  Yunzhi Zhuge \orcid{0000-0002-4288-4516} \and 
  Huchuan Lu \orcid{0000-0002-6668-9758}
}

\authorrunning{Y. Shan, Y. Zhuge, and H. Lu} 

\institute{
Yongqi Shan is with the School of Computer Science and Technology, 
Dalian University of Technology, Dalian 116081, China.
(Email: 11312327@mail.dlut.edu.cn). \\
Yunzhi Zhuge and Huchuan Lu are with the School of Information and Communication Engineering, 
Dalian University of Technology, Dalian 116081, China. 
(Email: zgyz@dlut.edu.cn; lhchuan@dlut.edu.cn). \\
Corresponding author: Yunzhi Zhuge (zgyz@dlut.edu.cn).
}

\date{Published in Visual Intelligence 4, Article 4 (2026).\\\url{https://doi.org/10.1007/s44267-026-00107-3}}

\maketitle

\begin{abstract}
Semi-supervised semantic segmentation, which aims to utilize large volumes of unlabeled images to achieve accurate segmentation results with fewer human annotations, has attracted increasing attention. Prior methods were primarily based on the assumption that the pseudo-labels generated by each branch were sufficiently reliable to supervise the remaining branches. However, there is a concern that the errors in pseudo-labels could accumulate during co-training, potentially resulting in suboptimal performance. To this end, we present a novel framework called heterogeneous dual-branch voting supervision (HDVS), which is designed to enhance the reliability of pseudo-labels and mitigate the issues arising from pseudo-labeling. Specifically, based on the cross-supervision framework, we introduce a voting mechanism that correlates pseudo-labels from heterogeneous branches to produce pseudo-labels with enhanced reliability. Concurrently, we employ a feature communication module to introduce perturbations at the feature level in each branch to maximize prediction diversity across the dual branches while maintaining convergence. Comprehensive evaluations of the proposed HDVS on two benchmark datasets (PASCAL VOC 2012 and Cityscapes) demonstrate its superiority to the state-of-the-art approaches. 


\end{abstract}

\section{Introduction}
Recent advancements in artificial intelligence have been driven by the development of deep neural networks and the availability of large-scale, well-annotated training datasets~\cite{dosovitskiy2020image, xie2021segformer, wang2022pvt, kirillov2023segment, xu2023pidnet, li2023mask}. However, collecting high-quality annotations, particularly densely pixel-wise labels required for semantic segmentation, demands considerable effort~\cite{ding2025mevis,ding2025mosev2,ding2023mose,ding2023mevis}. This challenge has spurred a growing trend in research towards semi-supervised semantic segmentation, which leverages a combination of a few labeled and many unlabeled images to enhance segmentation performance~\cite{souly2017semi}.

\begin{figure*}[t]
    \centering
    \includegraphics[width=1\linewidth]{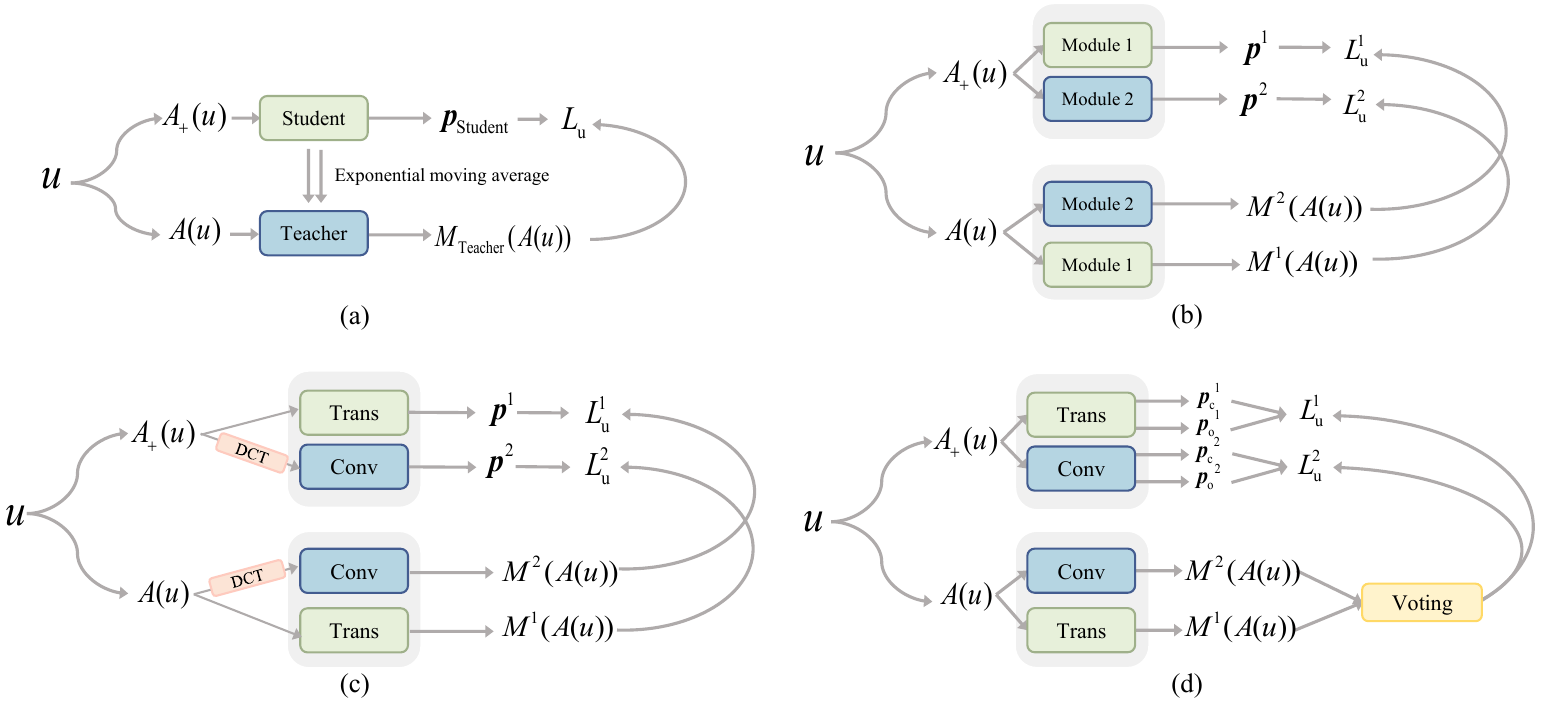}
    \caption{Overview of common semi-supervised frameworks. (a) FixMatch~\cite{sohn2020fixmatch} structure used in U$^2$PL~\cite{ wang2022semi}, (b) cross pseudo supervision (CPS)~\cite{chen2021semi}, (c) diverse co-training with discrete cosine transform (DCT) for image processing~\cite{li2023diverse}, and (d) our heterogeneous dual-branch voting supervision (HDVS) method where two models use voting to generate pseudo-labels. $u$ denotes an unlabeled image, $A$ and $A_+$ denote weak and strong data augmentations, $M^{1}$ and $M^{2}$ are the two segmentation models, and $M_{\mathrm{Teacher}}$ denotes the teacher model. Trans and Conv denote the Transformer and convolutional neural network (CNN) branches, $\bm{p}$ denotes their predictions, $\bm{p}_{\mathrm{o}}$ and $\bm{p}_{\mathrm{c}}$ denote the predictions from the original branch and the feature communication branch, respectively. $L_{\mathrm{u}}, L_{\mathrm{u}}^{1}, L_{\mathrm{u}}^{2}$ are the corresponding unsupervised losses.
}
    \label{fig:2}
\end{figure*}

In the realm of semi-supervised semantic segmentation, a significant research direction involves the use of teacher-student networks~\cite{berthelot2019remixmatch,berthelot2019mixmatch,sohn2020fixmatch,tarvainen2017mean,wang2022semi,chen2021semi,li2023diverse}. As illustrated in Fig.~\ref{fig:2}(a), methods such as U$^2$PL~\cite{wang2022semi} utilize a teacher network to segment unlabeled data and generate segmentation maps, from which pseudo-labels are derived. The parameters are continuously updated through the exponential moving average (EMA) derived from the student network. However, relying solely on a single network for pseudo-labels reconstruction will bring noise and incur confirmation biases~\cite{arazo2020pseudo,ouali2020overview}. The co-training pipeline, as proposed by Chen et al.~\cite{chen2021semi} and depicted in Fig.~\ref{fig:2}(b), introduces the cross pseudo supervision (CPS) strategy on two networks with identical architectures and different initializations to capture various characteristics for pseudo-labels. Subsequent advancements to CPS enhance its capabilities by introducing models with distinct initialization~\cite{fan2022ucc} and applying multiple perturbations~\cite{li2023cfcg} to expand the diversity of feature perturbation. Recently, Li et al.~\cite{li2023diverse} introduced a method to enhance diversity in co-training across input domains, augmentations, and architectures, showcasing the significant impact of diversity on achieving high-performance segmentation, with the architecture illustrated in Fig.~\ref{fig:2}(c). However, these methods have not fully exploited the deep collaboration among their diverse components, often resulting in unreliable pseudo-labels construction. As illustrated in Fig.~\ref{fig:2}(d), our key insight is to integrate complementary cues from heterogeneous networks through dynamic, confidence-aware voting, which exposes regions of consistently high/low confidence while letting each branch's inductive bias refine supervision.

\begin{figure}[t]
\centering
\scalebox{1}{
    \includegraphics[width=1\linewidth]{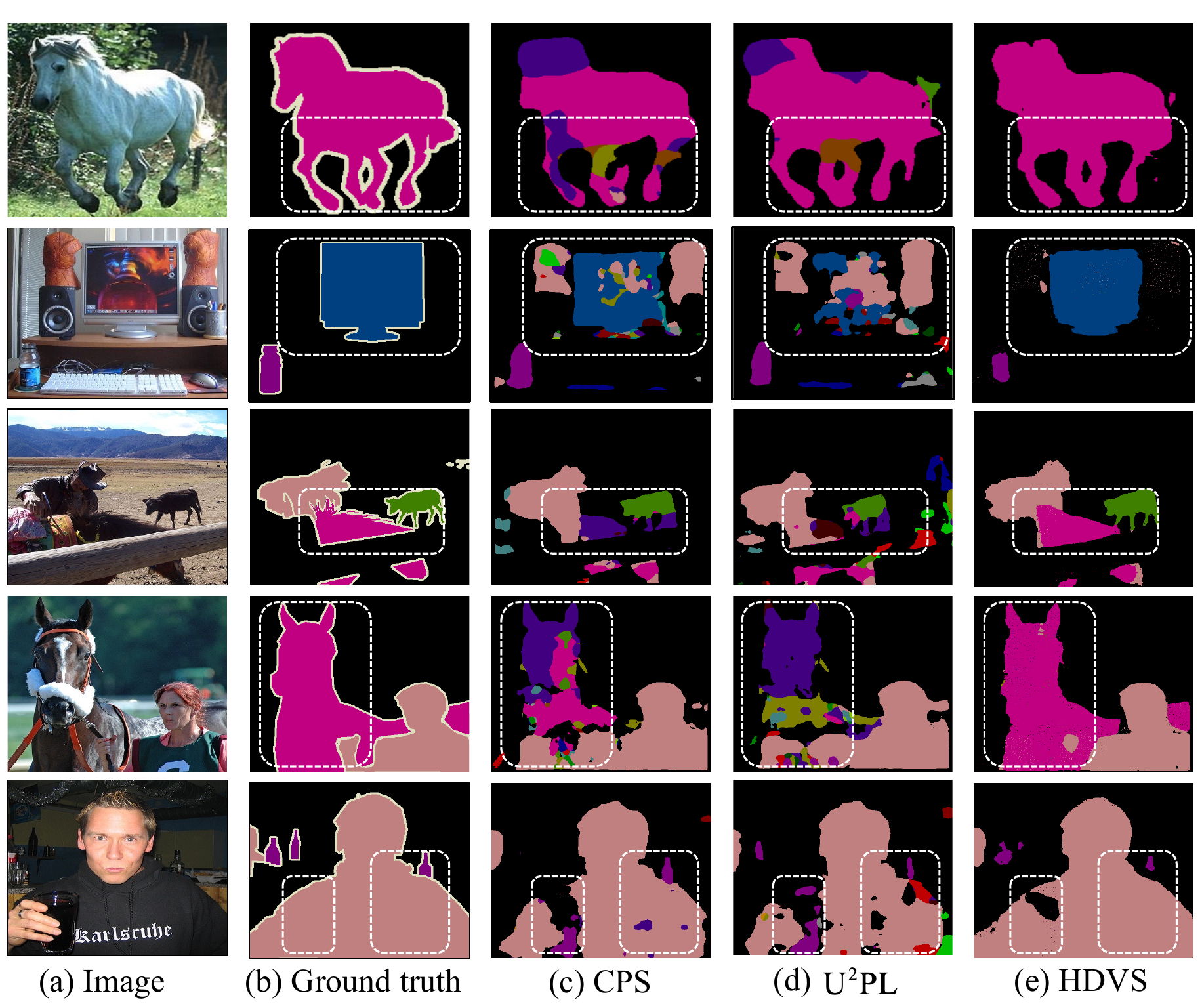}}
    \caption{Quality analysis of segmentation maps. From left to right: (a) input images, (b) manually annotated ground truth labels, (c) segmentation maps generated by CPS, (d) segmentation maps generated by U$^2$PL, and (e) segmentation maps produced by HDVS. All models are trained using the 1/8 classical partition protocol on the PASCAL VOC 2012 ~\cite{everingham2010pascal} train dataset.}
    \label{fig:1}
\end{figure}

The performance of semi-supervised semantic segmentation models significantly depends on the quality of the pseudo-labels, where high-quality pseudo-labels are typically characterized by minimal uncertainty~\cite{grandvalet2004semi}. Motivated by this insight, our approach aims to enhance collaborative training by integrating high-confidence predictions from heterogeneous models to produce superior pseudo-labels. We conduct a comparative visualization to highlight differences in the segmentation maps generated by CPS, U$^2$PL, and our method. As illustrated in Fig.~\ref{fig:1}, our model demonstrates robust performance across diverse challenges, including high-quality segmentation of complex structures such as the intricate leg regions of a horse (example in the first row), effective handling of complex backgrounds such as the decorations on either side of a computer (example in the second row), accurate detection of multiple small objects, such as people, cows, and horses (example in the third row), and superior performance in scenarios involving object overlap, such as between a person and a horse or between a person and a bottle (examples in the forth and fifth rows). This performance gain results from a virtuous cycle between the heterogeneous branches and the dynamic voting module. At each iteration, the branches produce diverse and complementary predictions, which are fused by the voting module into more reliable pseudo-labels. These pseudo-labels provide stronger supervision, further improving the branches' predictive capabilities. As training progresses, this process of prediction refinement, pseudo-labels enhancement, and model improvement continually elevates the quality of both pseudo-labels and segmentation outputs.

In this work, we argue that relying solely on a single network's predictions to generate pseudo-labels is suboptimal. Most prior methods either prioritize reliability via a single teacher--student network with confidence thresholds or pursue diversity via co-training/CPS pipeline (multi-view/augmentation/architecture) that swap per-branch argmax labels without explicit confidence fusion~\cite{chen2021semi,wang2022semi,li2023diverse}, thus risking confirmation bias. Instead, we advocate for an approach that harnesses collective wisdom and collaborative effort. Specifically, heterogeneous dual-branch voting supervision (HDVS) introduces a voting module after the decoders and couples it with a lightweight feature communication module (FCM): the voting module fuses the two outputs into a single pseudo-labels per pixel based on prediction confidences, whereas the FCM performs bidirectional, level-aligned feature communication to preserve global--local complementarity and maintain inter-branch diversity. Taken together, the FCM and the voting module operate within a single training loop: as feature communication proceeds between the heterogeneous branches, those same branches co-decide the labels via dynamic voting, jointly improving pseudo-labels reliability and preserving inter-branch diversity, thereby stabilizing training and yielding stronger segmentation performance.

Comprehensive experiments validate the efficacy of our HDVS, showcasing notable performance improvements of 0.7$\sim$2.9 mean intersection over union (mIoU) points across two public benchmarks, $\mathrm{i.e.,}$ PASCAL VOC 2012~\cite{everingham2010pascal} and Cityscapes~\cite{cordts2016cityscapes}. These results underscore the potential of HDVS as a scalable and foundational approach in semi-supervised segmentation. Our source code and model weights will be made publicly available.

\section{Related Work}
\subsection{Semi-Supervised Learning} 

Semi-supervised learning (SSL) aims to enhance model performance by using unlabeled data when labeled data is limited, and it has rapidly developed in recent years~\cite{grandvalet2004semi, chen2017semi, zhang2021flexmatch, wang2022freematch, sun2023semi}. The central challenge in SSL involves devising effective supervisory signals for a large volume of unlabeled data~\cite{oliver2018realistic,ding2025multimodal}. Current approaches in SSL primarily follow two foundational principles. The first principle is entropy minimization~\cite{grandvalet2004semi,lee2013pseudo,xie2020self,zoph2020rethinking,pham2021meta,ding2023decoupling}, which aims to boost model confidence by promoting predictions that minimize uncertainty on unlabeled data. Specifically, entropy minimization compels the model to deliver confident and decisive outputs, thereby reducing the entropy in its probability perturbations. This principle often complements self-training techniques~\cite{lee2013pseudo}, where high-confidence outputs, termed pseudo-labels, are generated for unlabeled data. These pseudo-labels, combined with manually labeled data, are then used to enhance the training process.

Another significant branch of research in SSL employs consistency regularization~\cite{jeong2019consistency,berthelot2019mixmatch,berthelot2019remixmatch,sohn2020fixmatch,rizve2021defense}, which compels models to produce consistent predictions across various perturbations of the identical input. CSD~\cite{jeong2019consistency} integrates consistency constraints into object classification and localization tasks while utilizing background elimination to reduce the influence of background elements on detection performance. MixMatch~\cite{berthelot2019mixmatch} combines data augmentation, low-entropy label guessing for unlabeled data, and a mixing strategy to enforce prediction consistency, which helps in smoothing prediction perturbations and regularizing the model. Building on the MixMatch, ReMixMatch~\cite{berthelot2019remixmatch} introduces perturbation alignment to improve class balance and augmentation anchoring, ensuring consistent predictions across strongly augmented versions of the same unlabeled instance. FixMatch~\cite{sohn2020fixmatch} uses pseudo-labels generated from weakly augmented data to guide the training on strongly augmented versions, only incorporating pseudo-labels that meet a confidence threshold. Expanding on FixMatch, UPS~\cite{rizve2021defense} enhances the training process by reducing noise interference, significantly increasing the accuracy of pseudo-labels, and addressing uncertainties related to both the model and the data.

\subsection{Semi-supervised Semantic Segmentation}
Integrating consistency learning with pseudo-labeling strategies has significantly advanced semi-supervised semantic segmentation~\cite{miyato2018virtual,ouali2020semi,xie2020unsupervised}. Following the consistency regularization theorem, French et al.~\cite{french2019semi} successfully applied the Cutout~\cite{devries2017improved} and CutMix~\cite{yun2019cutmix} techniques to semi-supervised semantic segmentation, demonstrating their effectiveness. CCT~\cite{ouali2020semi} employs cross-consistency training, which trains the model to align predictions across different augmentations or views of an image. CPS~\cite{chen2021semi} uses a strategy of which pseudo-labels generated by separate but architecturally identical networks are used for cross-training, facilitating mutual enhancement and error rectification within the networks. UDA~\cite{xie2020unsupervised} enhances the consistency training framework by replacing simple noise operations with more advanced data augmentation methods, such as RandAugment~\cite{cubuk2020randaugment} and back-translation~\cite{sennrich2015improving}, to improve the utilization of unlabeled data.

Another prominent research trend in semi-supervised semantic segmentation utilizes the teacher-student architecture, where the student progressively updates the teacher via EMA and then uses its generated pseudo-labels to supervise the student's training~\cite{sohn2020fixmatch, yuan2021simple, wang2022semi, liu2022perturbed, zhao2023augmentation}. On the one hand, various studies aim to augment pseudo-labels' utility by learning to assess their reliability with confidence scoring techniques. U$^2$PL~\cite{wang2022semi} designates unreliable pixels as negative samples within a contrastive learning framework. ELN~\cite{kwon2022semi} improves the quality of pseudo-labels through the integration of a trainable correction network. On the other hand, numerous studies emphasize the critical importance of data augmentation in this field~\cite{yuan2021simple,liu2022perturbed,zhao2023augmentation}. UniMatch~\cite{yang2023revisiting} explores innovative augmentation techniques for creating perturbed data views. AugSeg~\cite{zhao2023augmentation} innovatively enhances less confident unlabeled samples by leveraging confidently labeled samples through the application of the CutMix technique in a flexible manner. Contrary to common methods that rely on data augmentation or the design of additional correction networks, our approach employs a simple yet effective voting mechanism to enhance the quality of pseudo-labels. Building upon the foundational cross-supervision framework, our model innovates through the integration of a heterogeneous dual-network architecture complemented by a feature communication module and the voting mechanism.

\section{Method}

\subsection{Preliminary}\label{sec:preliminary}

Originally developed for semi-supervised classification tasks~\cite{qiao2018deep,ke2019dual}, the co-training approach has been successfully extended to semi-supervised semantic segmentation by CPS~\cite{chen2021semi}. As depicted in Fig.~\ref{fig:2}(b), this approach employs a dual-branch architecture, with each branch comprising an encoder and a decoder. These branches are collaboratively trained using a combination of a small set of labeled data, denoted as $B_{\mathrm{l}}$, and a larger pool of unlabeled data, denoted as $B_{\mathrm{u}}$. We denote the two segmentation models (branches) as $M^{0}$ and $M^{1}$. The training process is calculated as follows:
\begin{equation}
\mathcal{L} = \mathcal{L}_{\mathrm{s}} + \lambda \mathcal{L}_{\mathrm{u}},
\end{equation}
where $\mathcal{L}_{\mathrm{s}}$ signifies the supervised loss term, and $\mathcal{L}_{\mathrm{u}}$ represents the unsupervised loss term. $\lambda$ serves as a weighting coefficient to balance different losses. Without loss of generality, the supervised loss is formulated as follows:
\begin{equation}
\mathcal{L}_{\mathrm{s}} = \frac{1}{B_{\mathrm{l}} HW} \sum_{i=1}^{B_{\mathrm{l}}} \sum_{j=1}^{HW} \left( \ell_{\mathrm{ce}}(\bm{p}_{ij}^0, y^*_{ij}) + \ell_{\mathrm{ce}}(\bm{p}_{ij}^1, y^*_{ij}) \right),
\end{equation}
where $\ell_{\mathrm{ce}}$ denotes the pixel-wise cross-entropy loss applied to each pixel. $j$ represents the $j$-th pixel in the image or the corresponding segmentation mask of resolution $H \times W$, $\bm{p}_{ij}^0$ and $\bm{p}_{ij}^1$ are the predictions from the two network branches for the $j$-th pixel in the $i$-th image in the batch, and $y^*_{ij}$ is the ground truth label for that pixel.

The unsupervised loss employs a bidirectional cross-supervision mechanism, wherein the pixel-wise one-hot label from one network serves as the supervisory signal for the other, as delineated below: 
\begin{equation}
\mathcal{L}_{\mathrm{u}} = \frac{1}{B_{\mathrm{u}} HW} \sum_{i=1}^{B_{\mathrm{u}}}\sum_{j=1}^{HW}
\Bigl(\ell_{\mathrm{ce}}(\bm{p}_{ij}^0, \hat{y}^1_{ij})
+ \ell_{\mathrm{ce}}(\bm{p}_{ij}^1, \hat{y}^0_{ij})\Bigr),
\end{equation}
where $\hat{y}^0_{ij}$ and $\hat{y}^1_{ij}$ denote the pixel-wise one-hot pseudo-labels generated by the first and second networks, respectively.

\subsection{Heterogeneous Dual-Branch Network}\label{sec:hdbn}
To increase the diversity between models and enhance the co-supervision process, our model adopts a heterogeneous dual-branch structure. For each labeled image, the loss is directly computed by utilizing the manually labeled ground truth. Specifically, we initiate the process by subjecting the data to weak data augmentation. Subsequently, the augmented data is fed into the encoders of two distinct models, generating respective image feature maps. In the original branch, the features are directly input into the decoder within each model, yielding the prediction outcome $\bm{p}_{\mathrm{o}}$. Simultaneously, in the feature communication branch, features undergo communication through the FCM, incorporating features from the counterpart model. These communication features are fed into the decoder to generate the prediction outcome $\bm{p}_{\mathrm{c}}$. The specific details of the FCM will be elaborated in Section~\ref{sec:fcm}. Therefore, the supervised loss for our model is defined as follows:
\begin{equation}
\begin{split}
\mathcal{L}_{\mathrm{s}} = \sum_{\alpha=0}^1 \frac{1}{B_{\mathrm{l}} HW} \sum_{i=1}^{B_{\mathrm{l}}} \sum_{j=1}^{HW} &\left(\ell_{\mathrm{ce}}(\bm{p}_{\mathrm{o}ij}^\alpha, y_{ij}^*) \right.\\
&\left. + \ell_{\mathrm{ce}}(\bm{p}_{\mathrm{c}ij}^\alpha, y_{ij}^*)\right),
\end{split}
\end{equation}
where the superscript $\alpha$ denotes the segmentation networks, with $\alpha = 0$ corresponding to the Transformer model and $\alpha = 1$ to the convolutional neural network (CNN) model.

\begin{figure*}[t]
    \centering
    \includegraphics[width=1\linewidth]{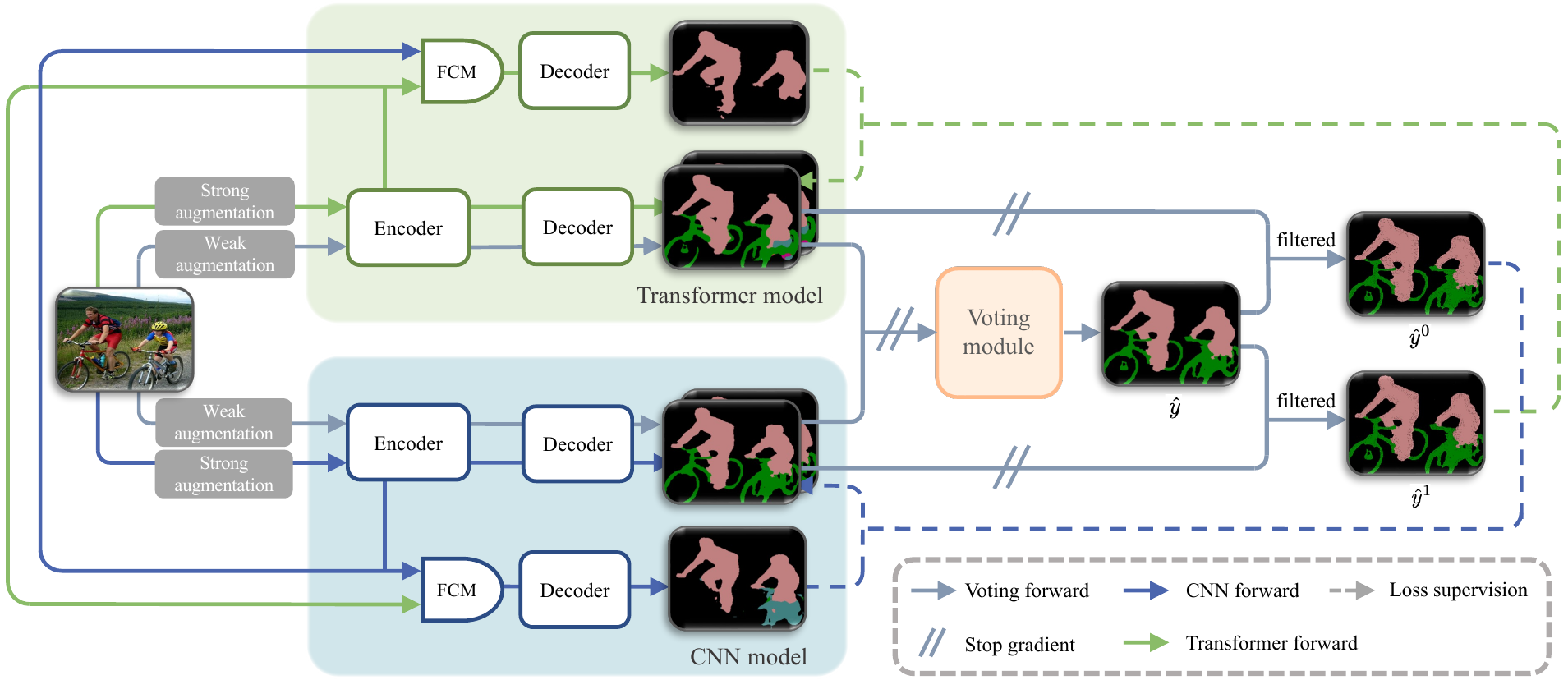}
    \caption{Main pipeline of our HDVS method. It integrates two distinct parallel architectures, $\mathrm{i.e.}$, CNNs and Transformers, to capture pseudo-labels from different perspectives. To improve label accuracy, a fusion and voting strategy is applied across these architectures. The feature communication module (FCM) on each branch supports implicit feature-level integration, while a specialized voting module explicitly refines the pseudo-labels generated by each branch. The refined labels, denoted as $\hat{y}$, and their filtered versions, $\hat{y}^0$ and $\hat{y}^1$, are used to jointly train both models, promoting synergistic learning across the architectures.}
    \label{fig:3}
    \vspace{-3mm}
\end{figure*}

For processing an unlabeled image, as illustrated in Fig.~\ref{fig:3}, the image first undergoes weak data augmentation before being input into two segmentation models. These models produce individual prediction results, denoted as $\bm{p}_{\mathrm{o}}^0$ and $\bm{p}_{\mathrm{o}}^1$. These results are then input into the voting module to generate pseudo-labels $\hat{y}$. The entropy of these predictions reflects the uncertainty of the model outputs, with higher entropy indicating lower confidence in the generated pseudo-labels. Following the method described in Ref.~\cite{wang2022semi}, we calculate the entropy based on $\bm{p}_{\mathrm{o}}^0$ and $\bm{p}_{\mathrm{o}}^1$. To enhance label reliability, we discard pseudo-labels associated with the top 20\% of pixel positions exhibiting the highest entropy, resulting in refined pseudo-labels $\hat{y}^0$ and $\hat{y}^1$. Detailed workings of the voting module are further explored in Section~\ref{sec:voting}. The entropy $\mathcal{H}$ is calculated as follows:
\begin{equation}
\mathcal{H}^\alpha\left(\bm{p}_{\mathrm{o}ij}^\alpha \right) = -\sum_{c=0}^{C-1} \bm{p}_{\mathrm{o}ij}^\alpha (c) \log \bm{p}_{\mathrm{o}ij}^\alpha(c),
\end{equation}
where $C$ denotes the total number of classes. Later, the image, subjected to strong data augmentation, is fed into the two models. Each model generates a prediction result on both the feature communication and the original branches, supervised by the filtered pseudo-labels. Therefore, our unsupervised loss is computed as
\begin{equation}
\begin{split}
\mathcal{L}_{\mathrm{u}} = \sum_{\alpha=0}^1 \frac{1}{B_{\mathrm{u}} H W} \sum_{i=1}^{B_{\mathrm{u}}} \sum_{j=1}^{HW} &\big( \ell_{\mathrm{ce}}(\bm{p}_{\mathrm{o}ij}^\alpha, \hat{y}_{ij}^{(1-\alpha)}) \\
&+ \ell_{\mathrm{ce}}(\bm{p}_{\mathrm{c}ij}^\alpha, \hat{y}_{ij}^{(1-\alpha)})\big).
\end{split}
\end{equation}

\subsection{Voting-based Pseudo-label Generation}\label{sec:voting}
The quality of pseudo-labels is fundamental for achieving high model accuracy. Noisy pseudo-labels can lead to the accumulation of inaccurate data and increase the risk of confirmation bias, as noted by Arazo et al.~\cite{arazo2020pseudo}. Most current semi-supervised semantic segmentation methods, such as the teacher-student structure~\cite{wang2022semi} and cross-supervision structure~\cite{chen2021semi}, rely on a single segmentation network to generate pseudo-labels. This dependence on a single perspective, which is inherently susceptible to noise, typically results in lower-quality pseudo-labels. Additionally, pseudo-labels of higher quality are generally associated with greater prediction confidence. To mitigate these challenges, we introduce a voting module that evaluates prediction confidence across dual network branches. This module enables the selective incorporation of higher-quality pseudo-labels into the training process, thereby enhancing overall model performance.

The process for processing an unlabeled image $u_i$ is detailed in Fig.~\ref{fig:3}. Initially, the image undergoes weak data augmentation and is then input into two separate models, which generate prediction vectors ${\bm{p}_{\mathrm{o}i}^0, \bm{p}_{\mathrm{o}i}^1} \in \mathbb{R}^{H \times W \times C}$. Each vector's highest confidence score is identified, and the corresponding probability is obtained via a $\text{softmax}$ operation. Following this, a random variable, $r$, uniformly distributed between 0 and 1, is generated to aid the decision-making process. This variable is compared with the normalized probabilities of the most confidently predicted category from each model. The result of this comparison determines which model's output will be used as the final pseudo-labels. By employing this method, we effectively implement a soft voting mechanism that adaptively integrates predictions from the heterogeneous models while introducing an element of uncertainty into the decision-making process. The final pseudo-labels are determined based on this comparative assessment as
\vspace{-1mm}
\begin{equation}
\ \mathrm{prob}_{ij}^0, \mathrm{label}_{ij}^0= \text{torch.max}(\text{softmax}(\bm{p}_{\mathrm{o}ij}^0)),
\label{eq:7}
\end{equation}
\vspace{-7mm}
\begin{equation}
\ \mathrm{prob}_{ij}^1, \mathrm{label}_{ij}^1=\text{torch.max}(\text{softmax}(\bm{p}_{\mathrm{o}ij}^1)),
\label{eq:8}
\end{equation}
\vspace{-7mm}
\begin{equation}
\hat{y}_{ij}=\begin{cases}
 \mathrm{label}_{ij}^0, & \text{ if } r<\dfrac{\mathrm{prob}_{ij}^0}{\mathrm{prob}_{ij}^0+\mathrm{prob}_{ij}^1}, \\
 \mathrm{label}_{ij}^1, & \text{else,} 
\end{cases}
\label{eq:9}
\end{equation}
where $\mathrm{prob}_{ij}^0$ and $\mathrm{prob}_{ij}^1 \subseteq \left [ 0,1 \right )$ denote the probabilities of the most confident classes by the two segmentation networks, determined using torch.max, which returns the maximum probability and its corresponding class index along the class dimension. These probabilities correspond to the pseudo-labels $\mathrm{label}_{ij}^0$ and $\mathrm{label}_{ij}^1$. Additionally, the variable $\hat{y}_{ij} \in \{0, 1,\ldots, C-1\}$ signifies the high-quality pseudo-labels generated through the voting module. The specific procedural steps of this voting mechanism are detailed in Algorithm~\ref{Algorithm1}.

\SetAlgoCaptionSeparator{}
\SetKwHangingKw{KwIn}{Input}
\SetKwHangingKw{KwOut}{Output}
\SetKwFor{For}{for}{do}{end~for}
\SetAlgoNoLine
\begin{algorithm}[htpb]
\caption{Voting-based pseudo-label generation pipeline}
\label{Algorithm1}
\DontPrintSemicolon

\KwIn{Sample $N$, unlabeled images $B_{\mathrm{u}}$}
\KwOut{Voting-based pseudo-labels $\hat{y} \in \mathbb{R}^{N\times H \times W}$}

\For{$u_i \in B_{\mathrm{u}}$}{
    \textcolor[rgb]{0.514,0.725,0.620}{\# Perform weak data augmentation $A_{\mathrm{w}}$.}
    
    $x_i \leftarrow A_{\mathrm{w}}(u_i)$;
    
    \textcolor[rgb]{0.514,0.725,0.620}{\# Get predictions from the two models $M^0$ and $M^1$.}
    
    $\bm{p}_{\mathrm{o}i}^0 \leftarrow M^0(x_i)$, 
    $\bm{p}_{\mathrm{o}i}^1 \leftarrow M^1(x_i)$;

    \textcolor[rgb]{0.514,0.725,0.620}{\# Sample a uniform random number.}
    
    $r \sim \mathcal{U}(0,1)$;
    
    \For{$j \in H\times W$}{
        \textcolor[rgb]{0.514,0.725,0.620}{\# Get confidence scores.} 
        $\mathrm{pred}_{ij}^0 \leftarrow \text{softmax}(\bm{p}_{\mathrm{o}ij}^0)$;
        $\mathrm{pred}_{ij}^1 \leftarrow \text{softmax}(\bm{p}_{\mathrm{o}ij}^1)$;
        
        \textcolor[rgb]{0.514,0.725,0.620}{\# Get the top-1 category (argmax) and its corresponding probability.}
        $\mathrm{prob}_{ij}^0, \mathrm{label}_{ij}^0\leftarrow \text{torch.max}(\mathrm{pred}_{ij}^0)$;
        $\mathrm{prob}_{ij}^1, \mathrm{label}_{ij}^1 \leftarrow \text{torch.max}(\mathrm{pred}_{ij}^1)$;
        
        \textcolor[rgb]{0.514,0.725,0.620}{\# Obtaining pseudo-labels through voting.}
        $\begin{aligned}
        \hat{y}_{ij}=\begin{cases}
        \mathrm{label}_{ij}^0, & \text{ if } r<\frac{\mathrm{prob}_{ij}^0}{\mathrm{prob}_{ij}^0+\mathrm{prob}_{ij}^1}, \\
        \mathrm{label}_{ij}^1, & \text{else}. 
        \end{cases}
        \end{aligned}$ 
    }
}
\end{algorithm}

\subsection{Feature Communication Module}\label{sec:fcm}
Perturbations at the feature level have been widely applied in semi-supervised semantic segmentation to enhance model performance, as demonstrated in previous works~\cite{liu2022perturbed,yang2023revisiting}. Despite their effectiveness, the full potential of the feature space remains underutilized due to the limitations of features derived from a single model encoder. To address this, we developed the feature communication module FCM for each branch of our model. This module facilitates the integration of complementary information from different model encoders and optimizes feature communication. Additionally, it introduces controlled perturbations to enrich feature interactions, thereby enhancing the robustness and accuracy of our method.

\begin{figure}
    \centering
    \scalebox{0.9}{
    \includegraphics[width=1\linewidth]{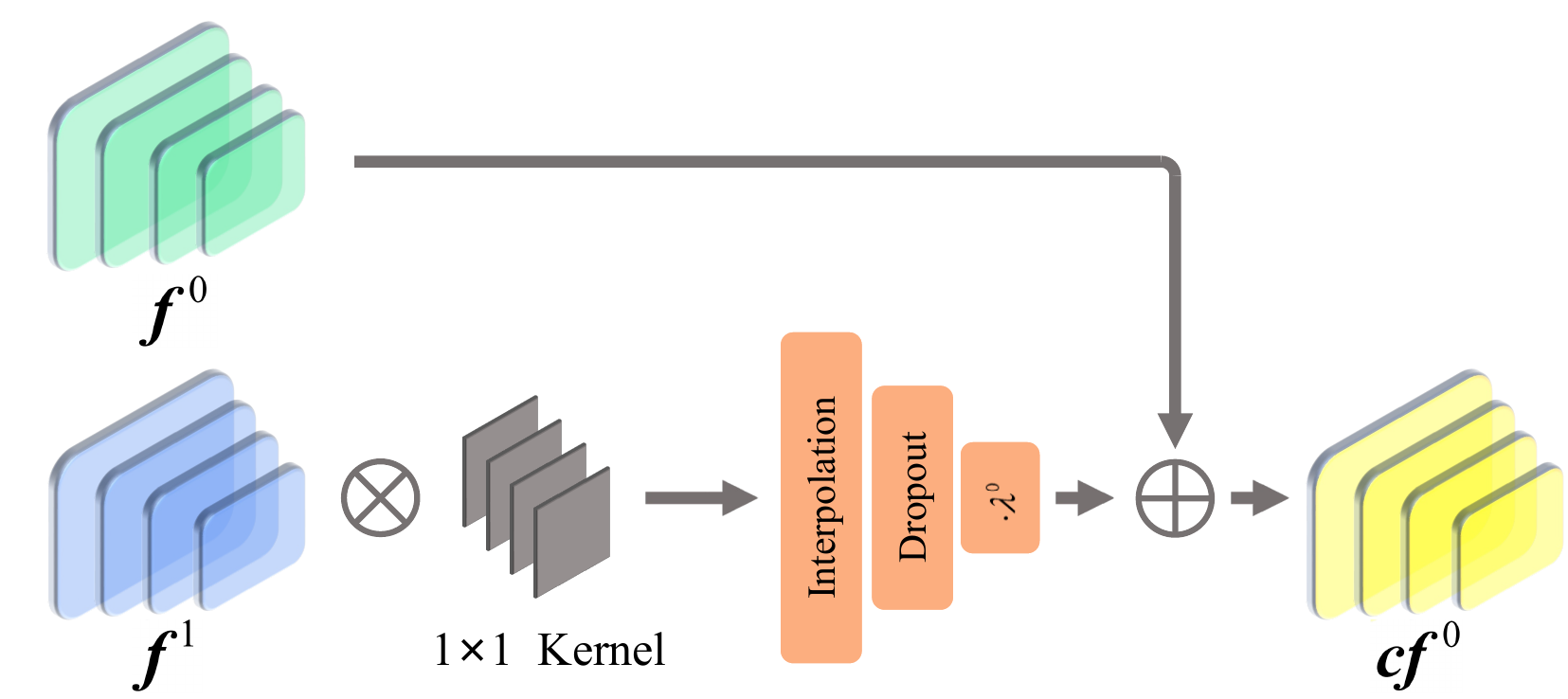}}
    \caption{
   Pipeline of the FCM in the Transformer branch. This module is designed to facilitate feature communication between two heterogeneous segmentation models while introducing controlled perturbations. The FCM effectively aligns and integrates features from both models to enhance segmentation performance. $\bm{f}^{0}$ and $\bm{f}^{1}$ denote features from the two encoders, $\bm{cf}^{0}$ is the communicated feature in the Transformer branch, and $\lambda^{0}$ controls the strength of injected features. \raisebox{-0.25ex}{\includegraphics[width=0.015\textwidth]{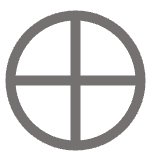}} denotes feature convolution. \raisebox{-0.25ex}{\includegraphics[width=0.015\textwidth]{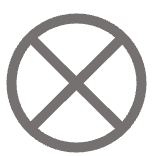}} denotes feature summation.}
    \label{fig:4}
    \vspace{-3mm}
\end{figure}

Taking the Transformer branch as an example, as depicted in Fig.~\ref{fig:4}, the features $\bm{f}^1$ extracted from the CNN encoder are integrated with the features $\bm{f}^0$ extracted from the Transformer encoder using the FCM. The multi-level features of $\bm{f}^0$ and $\bm{f}^1$ are then combined through a series of operations, including $1\times1$ convolutions, interpolation, dropout, and scaling coefficients. These operations ensure that both feature sets align in width, height, and channel dimensions at each level. This integration process can be represented as follows:
\vspace{-1mm}
\begin{equation}
\bm{cf}_{i}^0 = \bm{f}_{i}^0 + \lambda^0 \cdot \text{Dropout}\left(\mathcal{P}^0\left(\bm{f}_{i}^1\right)\right),
\end{equation}
\vspace{-7mm}
\begin{equation} 
\bm{cf}_{i}^1=\bm{f}_{i}^1+ \lambda^1 \cdot  \text{Dropout}\left ( \mathcal{P}^1\left (\bm{f}_{i}^0 \right )  \right ), 
\end{equation}
where $\bm{cf}_{i}^0$ and $\bm{cf}_{i}^1$ represent the features after undergoing processing by the FCM in their respective models, $\mathcal{P}^0$ and $\mathcal{P}^1$ represent adjustments made by the modules through interpolation and convolution operations, and $\lambda^0$ and $\lambda^1$ act as scaling coefficients for feature communication. \text{Dropout} is used for regularization by randomly dropping communicated features during training.

\subsection{Overall Pipeline of HDVS}\label{sec:HDVS}
HDVS integrates the FCM and a dynamic voting module, and jointly trains two heterogeneous branches. It is trained on a small labeled set $B_{\mathrm{l}}$ and a large unlabeled set $B_{\mathrm{u}}$. For each labeled image, the supervised loss $\mathcal{L}_{\mathrm{s}}$ is computed as the pixel-wise cross-entropy between each branch's prediction and the ground truth. For each unlabeled image, as shown in Fig.~\ref{fig:3}, we first apply weak augmentation $A_{\mathrm{w}}$ to obtain predictions from both branches. A dynamic voting module then fuses these predictions into pseudo-labels $\hat{y}$, followed by entropy-based filtering that masks the top $20\%$ most uncertain pixels to produce $\hat{y}^0$ and $\hat{y}^1$. Next, we apply strong augmentation $A_{\mathrm{s}}$ and forward both the original branch and the feature communication branch to obtain $\bm{p}_{\mathrm{o}}^\alpha$ and $\bm{p}_{\mathrm{c}}^\alpha$ ($\alpha\!\in\!\{0,1\}$), where the FCM injects controlled, cross-branch feature communication to preserve prediction diversity while stabilizing convergence. Finally, the branches are trained under cross-supervision against the filtered pseudo-labels to form the unsupervised loss $\mathcal{L}_{\mathrm{u}}$, with the overall objective $\mathcal{L}=\mathcal{L}_{\mathrm{s}}+\lambda\,\mathcal{L}_{\mathrm{u}}$.

\section{Experiment}
\subsection{Experiment Setup}

\textbf{Datasets.}
The PASCAL VOC 2012 dataset~\cite{everingham2010pascal} serves as a widely recognized benchmark for semantic segmentation, featuring 20 object classes plus a background class. It includes 1464 images in the training set and 1449 images in the validation set. 
In line with prior works~\cite{chen2021semi, hu2021semi, yang2022st++, wang2022semi, zhao2023augmentation}, we adopt two common training protocols: 
(1) VOC Train, a classic partition that uses only the 1464 high-quality labeled images from the VOC 2012 train set; and 
(2) VOC Aug, an augmented partition that expands the training data to 10582 images by including an additional 9118 images with weaker annotations from the semantic boundaries dataset (SBD)~\cite{hariharan2011semantic}.

The Cityscapes dataset~\cite{cordts2016cityscapes}, a widely recognized benchmark for semantic segmentation, focuses on urban scene understanding across 19 semantic classes. It includes 2975 images in the training set and 500 images in the validation set, each annotated at the pixel level to enable precise object delineation in complex city environments. Every high-resolution 1024$\times$2048 image spans a diverse range of urban settings from various cities. 

\textbf{Evaluation.}
To assess segmentation performance, we utilize the mean intersection over union termed as mIoU metric. Consistent with protocols described in Refs.~\cite{chen2021semi,wang2022semi}, we evaluate our model on the PASCAL VOC 2012 validation set by center-cropping images to a resolution of $513\times513$. For the Cityscapes dataset, we apply a sliding window technique to the validation set images, which maintain a resolution of $1024\times2048$, to thoroughly assess the model's performance across the entire image area.

\textbf{Implementation details.}
Our approach integrates two segmentation model architectures: the first is derived from Deeplabv3+~\cite{chen2018encoder}, utilizing a ResNet~\cite{he2016deep} backbone, while the second employs the SegFormer~\cite{xie2021segformer} model. Specifically, we conduct experiments using ResNet-50~\cite{he2016deep} with Mit-B2~\cite{xie2021segformer} or ResNet-101~\cite{he2016deep} with Mit-B3~\cite{xie2021segformer}. In our experiments, we find the CNN branch less tolerant to perturbations than the Transformer branch. Thus, we set $\lambda^0$ to 0.10 and $\lambda^1$ to 0.01 as scaling coefficients for the FCM in the Transformer and CNN models, respectively. Both FCMs utilize a dropout probability of 0.5. Following Refs.~\cite{chen2021semi,wang2022semi}, we incorporate weak data augmentation techniques such as flips and random cropping and apply the CutMix technique~\cite{yun2019cutmix} for strong data augmentation. We choose stochastic gradient descent (SGD) as the optimizer, complemented by synchronized batch normalization (Sync-BN). 

For the PASCAL VOC 2012 dataset, we configure the initial learning rate at 0.001, weight decay at 0.0001, crop size to 513×513, batch size to 4, and set the training duration to 150 epochs. For the Cityscapes dataset, we utilize online hard example mining cross-entropy (OHEM) loss~\cite{shrivastava2016training}, with the initial learning rate adjusted to 0.00075, weight decay at 0.0005, batch size at 2, and extend the training to 300 epochs. Consistent with approaches described in Refs.~\cite{wang2022semi,chen2021semi}, we employ a poly learning rate policy where the initial learning rate is scaled by $\left(1-\frac{\text{iter}}{\text{max\_iter}}\right)^{0.9}$, where $\text{iter}$ denotes the current training iteration and $\text{max\_iter}$ is the total number of training iterations.

\vspace{-5mm}
\subsection{Comparison with State-of-the-art Methods}

\textbf{PASCAL VOC 2012.}
\begin{table*}[t]
\caption{Comparison of mean intersection over union (mIoU) (\%) with state-of-the-art (SOTA) methods on the PASCAL VOC 2012\cite{everingham2010pascal} validation set with different backbones under various proportion protocols. ``Supervised'' denotes the supervised baseline without using unlabeled training data. The results of U$^2$PL~\cite{wang2022semi} (marked with $\dagger$) are obtained from Refs.~\cite{yang2023revisiting}, and it follows the same partition protocol as ours. ``VOC Train'' refers to using only the 1464 finely annotated images from the train set, ``VOC Aug'' includes an additional 9118 weakly labeled images from the semantic boundaries dataset~\cite{hariharan2011semantic}. Bold numbers indicate the best performance under each setting.}
\label{tab:table1}
\centering
\begin{tabular}{
  @{} >{\raggedright\arraybackslash}m{2.8cm} | 
  >{\centering\arraybackslash}m{1.3cm} 
  >{\centering\arraybackslash}m{1.3cm} 
  >{\centering\arraybackslash}m{1.3cm} 
  >{\centering\arraybackslash}m{1.3cm} 
  >{\centering\arraybackslash}m{1.3cm} | 
  >{\centering\arraybackslash}m{1.3cm} 
  >{\centering\arraybackslash}m{1.3cm} 
  >{\centering\arraybackslash}m{1.3cm} 
  >{\centering\arraybackslash}m{1.3cm} 
}

\hline
\multirow{3}{*}{Method}  & \multicolumn{5}{c|}{VOC Train} & \multicolumn{3}{c}{VOC Aug} \\
\cmidrule{2-9}
& 1/16 (92) & 1/8 (183) & 1/4 (366) & 1/2 (732) & Full (1464) & 1/16 (662) & 1/8 (1323) & 1/4 (2646) \\
\hline
\multicolumn{5}{l}{\textcolor[RGB]{90,90,90}{Results of using ResNet-50~\cite{he2016deep} as backbone}} \\
Supervised & 44.0& 52.3& 61.7& 65.9& 72.9 & 60.7&  67.7&  71.9\\
U$^2$PL$\dagger$  \cite{wang2022semi}& - & - & - & - & - & 72.0& 75.1& 76.2 \\
UniMatch~\cite{yang2023revisiting}& 71.9& 72.5& 76.0& 77.4& 78.7 & 75.8&  76.9& 76.8\\
AugSeg~\cite{zhao2023augmentation}& 64.2&72.2&  76.2&  77.4& 78.8 & 74.7& 76.0& 77.1 \\
DAW~\cite{sun2023daw}& 68.5& 73.1& 76.3& 78.6& 79.7& 76.2& 77.6& 77.4\\
Switch~\cite{na2023switching}&  70.7& 74.5& 76.4& 77.6& 78.1& -& -& -\\
Co-training~\cite{li2023diverse}& \textbf{73.1}& 74.7& 77.1& 78.8&80.2 & 76.2& 77.0& 77.5\\
Ours & 69.9& \textbf{75.6}& \textbf{79.6}& \textbf{81.3}&\textbf{83.1}&\textbf{78.2}& \textbf{79.3}& \textbf{79.9}\\
\hline
\multicolumn{5}{l}{\textcolor[RGB]{90,90,90}{Results of using ResNet-101~\cite{he2016deep} as backbone}} \\
Supervised&  43.9& 59.1& 65.9& 70.9&75.0&66.6&  70.5& 74.5 \\
CPS \cite{chen2021semi}&  64.1& 67.4& 71.7& 75.9 & -& 72.2& 75.8& 77.6 \\
U$^2$PL$\dagger$ \cite{wang2022semi}& 68.0& 69.2& 73.7& 76.2&79.5& 74.4&  77.6& 78.7 \\
UniMatch~\cite{yang2023revisiting}& 75.2& 77.2& 78.8& 79.9& 81.2& 78.1& 78.4& 79.2 \\
AugSeg~\cite{zhao2023augmentation}& 71.1& 75.5&  78.8&  80.3& 81.4 &77.0& 77.3& 78.8 \\
DAW~\cite{sun2023daw}&  75.5& 77.6& 79.8& 80.7& 82.2& 78.9& 79.2& 80.0\\
Co-training \cite{li2023diverse}&  75.7& 77.7& 80.1& 80.9& 82.0 & 77.6& 79.0& 80.0\\
AllSpark~\cite{wang2024allspark}&  76.1& 78.4& 79.8& 80.8& 82.1 & 78.3& 80.0 & 80.4\\
CorrMatch~\cite{sun2024corrmatch}&  \textbf{76.4}& 78.5& 79.4& 80.6& 81.8 & 78.4& 79.3 & 79.6\\
Ours & 72.9& \textbf{79.4}&  \textbf{81.2}&  \textbf{82.9}& \textbf{83.8}& \textbf{79.1}& \textbf{80.7}&\textbf{81.0}\\
\hline
\end{tabular}
\vspace{-3mm}
\end{table*}

Table~\ref{tab:table1} presents the performance of our HDVS method under both classic and blended partitioning schemes, in comparison with benchmarks established by previous state-of-the-art (SOTA) methods. The results demonstrate that HDVS, regardless of using ResNet-50 or ResNet-101 backbones, consistently delivers robust performance across different amounts of labeled data. Notably, our method surpasses the current SOTA method, diverse co-training~\cite{li2023diverse}, which also employs a combination of Transformer and CNN architectures. HDVS outperforms this method in four out of five categories under classic settings and in all categories under blended settings. Specifically, it achieves an average accuracy improvement of 1.5\% with the ResNet-50 backbone and 1.0\% with the ResNet-101 backbone.

A critical analysis of these results highlights the pivotal role of pseudo-labels quality. For instance, employing ResNet-50, the performance with 2646 coarsely annotated images lags by 1.4\% compared to using 732 finely annotated images, despite the fourfold increase in training data quantity. This phenomenon emphasizes the value of high-quality annotations. Additionally, it showcases the considerable effectiveness of our voting mechanism in producing high-quality pseudo-labels, significantly enhancing the overall model's performance.

\textbf{Cityscapes.} Table~\ref{tab:table2} presents the results achieved with ResNet-50 as the backbone on the Cityscapes train set, spanning 1/16, 1/8, 1/4, and 1/2 data partitions. Our HDVS achieves competitive results under different partition schemes. Additionally, it surpasses the previous SOTA methods by 0.6\% and 0.4\% mIoU under 1/8 and 1/2 partition protocols, respectively. 

\begin{table}[t]
\caption{Comparison of mIoU (\%) with SOTA methods on the Cityscapes~\cite{cordts2016cityscapes} validation set with ResNet-50 backbone under various partition protocols.}
\vspace{-3mm}
\scalebox{0.9}{
        \label{tab:table2}
        \centering
            \begin{tabular}[t]{@{}l>{\centering\arraybackslash}p{1.15cm}>{\centering\arraybackslash}p{1.15cm}>{\centering\arraybackslash}p{1.15cm}>{\centering\arraybackslash}p{1.15cm}}
            \hline
                 &  1/16 (186)&  1/8 (372)&  1/4 (744)& 1/2 (1488)\\
                 \hline
                Supervised       & 63.3& 68.7& 74.1& 76.6\\
              CPS \cite{chen2021semi}& 69.8& 74.4& 76.9& 78.6\\
              U$^2$PL \cite{wang2022semi}& 69.0& 73.0& 76.3& 77.2\\
              AugSeg \cite{zhao2023augmentation}& 73.7& 76.5& \textbf{78.8}& 79.3\\
              
             Co-training \cite{li2023diverse}& - & 76.3& 77.1& - \\
             DAW \cite{sun2023daw}& \textbf{75.2} & 76.3& 77.7& 79.2 \\
             UniMatch \cite{yang2023revisiting}& 75.0& 76.8& 77.5& 78.6\\
             Ours&  73.1&  \textbf{76.9}&  77.8& \textbf{79.7}\\
              \hline
              \vspace{-8mm}
            \end{tabular}
            }
\end{table}

\begin{figure*}[htpb!]
    \centering
    \scalebox{1}{
    \includegraphics[width=1\linewidth]{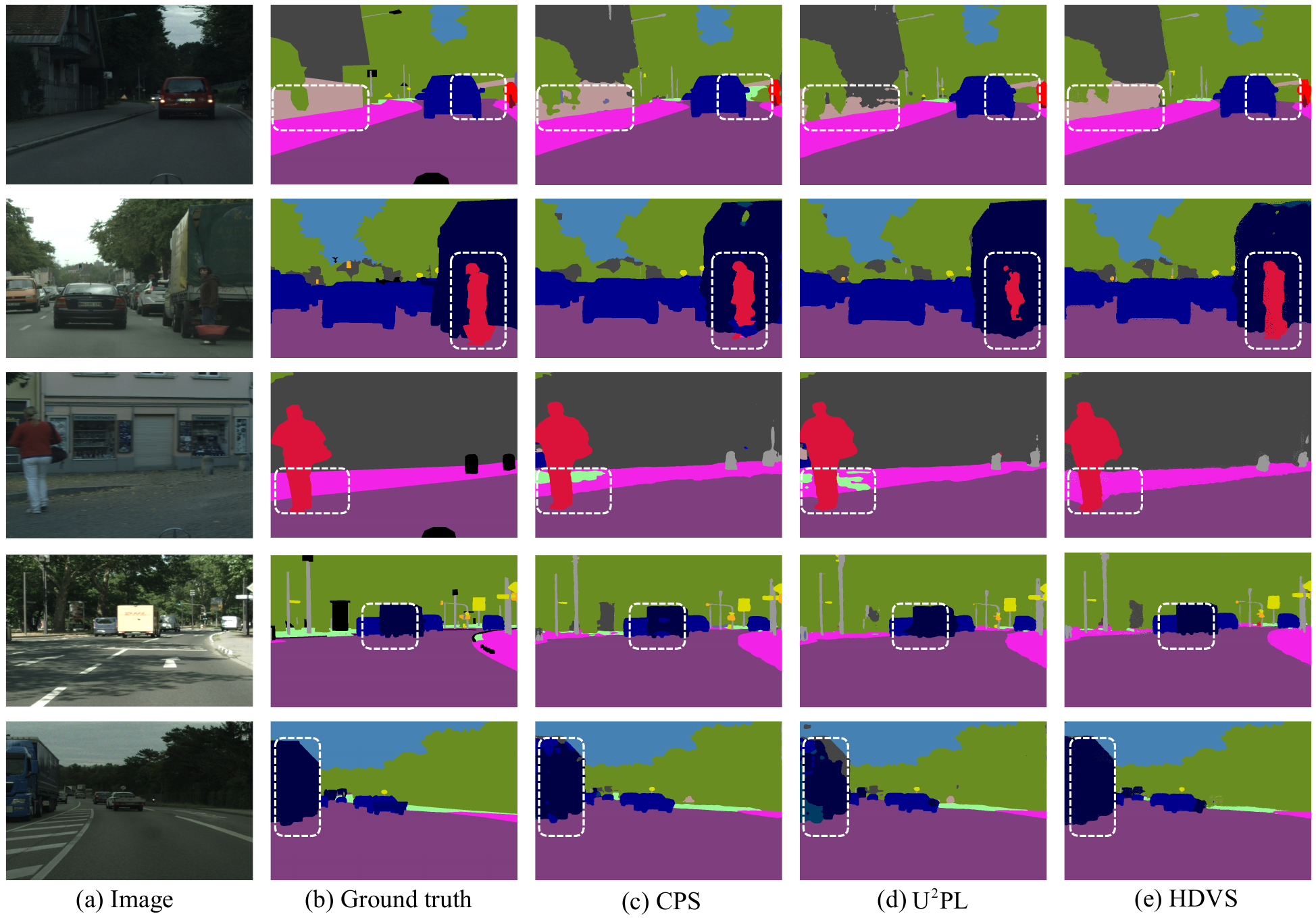}}
    \vspace{-4mm}
    \caption{Qualitative results on the Cityscapes validation~\cite{cordts2016cityscapes} set. From left to right, the categories are (a) input images, (b) ground truth labels annotated manually, (c) segmentation maps created by CPS, (d) segmentation maps produced by U$^2$PL, and (e) segmentation maps generated by our HDVS framework. All models are trained following the 1/2 (1488) partition protocol on the Cityscapes train set.}
    \label{fig:figure5}
    \vspace{-5mm}
\end{figure*}

A visual analysis of segmentation maps is presented in Fig.~\ref{fig:figure5}, which compares segmentation maps generated by CPS, U$^2$PL, and our HDVS on the Cityscapes validation set. All three models, utilizing ResNet-50 as the backbone, are trained using the Cityscapes train set under the 1/2 (1488) partition protocol. The results clearly show that the HDVS model performs exceptionally well in various challenging scenes. As illustrated in Fig.~\ref{fig:figure5}, in the first row, HDVS demonstrates remarkable capability in precisely capturing fine scene details. In the second and third rows, when confronted with scenes where the target is surrounded by complex environments, HDVS accurately delineates both the target boundaries and the surrounding background. The fourth row highlights its superior performance in scenarios involving overlapping objects, where HDVS successfully outlines complex structures such as trucks. Finally, the fifth row showcases HDVS's strong proficiency in handling scenes densely populated with numerous small objects, underscoring its exceptional effectiveness across diverse and complex urban environments.

\subsection{Feature Visualizations}
\begin{figure}[t]
    \centering
    \scalebox{1}{\includegraphics[width=1\linewidth]{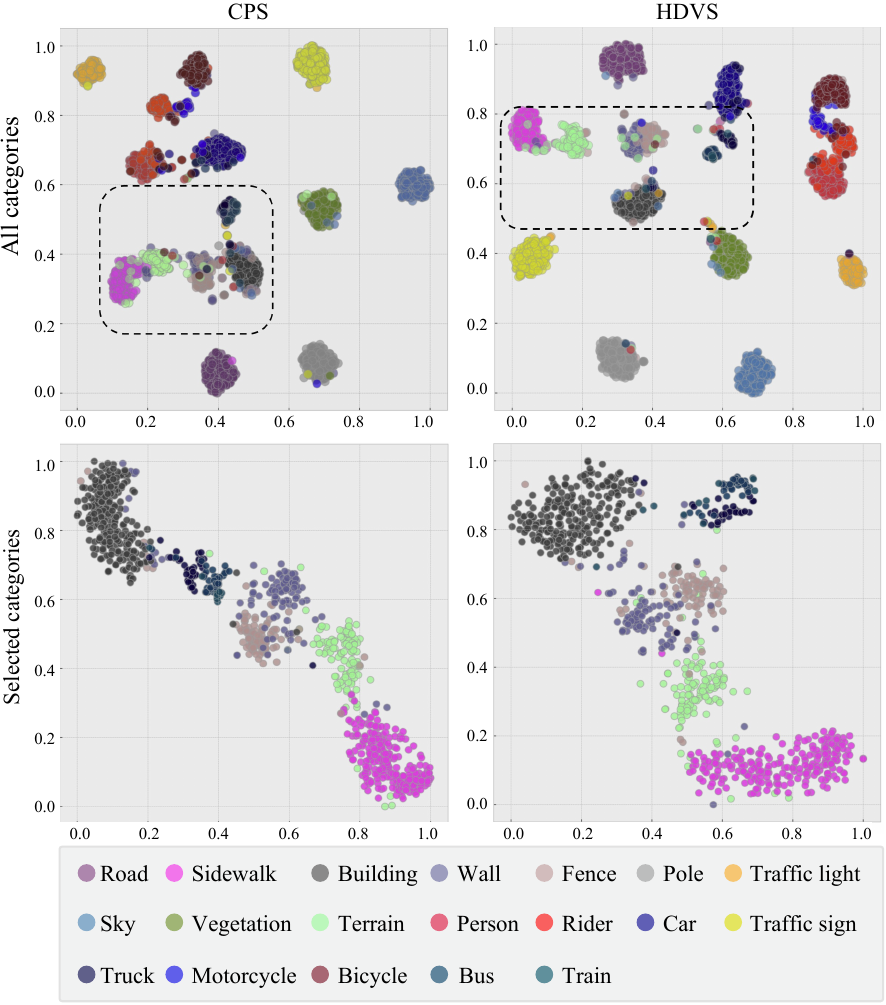}}
    \caption{$t$-SNE~\cite{maaten2008visualizing} visualization of extracted features by CPS and HDVS. Both models are trained on the Cityscapes train set following the 1/2 (1488) partition protocol, and the features are extracted from the Cityscapes validation set for display. The first row shows the visualization results for all categories, while the second row presents the visualization analysis for eight selected categories. The first and second columns display the visualization results for the CPS and HDVS models, respectively.}
    \label{fig:figure6}
    \vspace{-5mm}
\end{figure}

In Fig.~\ref{fig:figure6}, we present a $t$-SNE~\cite{maaten2008visualizing} visualization analysis of features extracted by CPS and our HDVS method, both using ResNet-50 as the backbone and trained on the Cityscapes training dataset with the 1/2 (1488) partition protocol. The first row shows the visualization results for all 19 categories in the Cityscapes dataset. These visualizations indicate that the HDVS model achieves superior class separation compared to CPS. Specifically, within regions outlined by the black dashed lines, CPS features exhibit significant overlap among eight categories: terrain, sidewalk, fence, wall, building, truck, bus, and train, indicating potential feature confusion. In contrast, our HDVS effectively distinguishes these categories, demonstrating exceptional precision in feature extraction and category discrimination. The second row focuses on these eight categories, further highlighting the differences. CPS features tend to converge, causing overlapping feature centers, whereas HDVS shows well-defined and dispersed feature centers for each category. This distinct separation underscores the enhanced capability of our method in discriminating features across multiple categories, showcasing its improved performance in feature discrimination.

\begin{table*}[t]
    \caption{Ablation study on the effectiveness of heterogeneous dual-branch voting supervision components. CPS$^*$ replicates CPS~\cite{chen2021semi} with aligned pseudo-label filtering and hyperparameters. Res-/Seg-CPS$^*$ denote dual convolutional neural network/ dual Transformer baselines; H-CPS$^*$ uses a heterogeneous CNN+Transformer pair. C-CPS$^*$ augments the model with the FCM and its communication branch, while V-CPS$^*$ adds the voting module for pseudo-label generation. HDVS combines both the feature communication and voting module. All results use a ResNet-50 backbone, and Mit-B2~\cite{xie2021segformer} is paired with ResNet-50 in the heterogeneous setting. $\checkmark$ and ``-'' indicate that the corresponding component is enabled and disabled, respectively. FCM denotes the feature communication module.}
    \vspace{1mm}
    \centering
    \renewcommand{\arraystretch}{1.4}
    \begin{tabular}{lcccc|cc}
    \hline
        &\multicolumn{4}{c}{Configurations}& \multicolumn{2}{c}{mIoU}\\
        \hline
         &  Heterogeneous networks&  Voting module&  FCM&  Communication branch&   1464&1323\\
          \hline
         Res-CPS$^*$&  -&  -&  -&  -&  78.8&76.9\\
         Seg-CPS$^*$&  -&  -&  -&  -&  80.9&77.8\\
         Res-C-CPS$^*$&  -&  -&  \checkmark& \checkmark &  78.9& 76.6\\
         Seg-C-CPS$^*$&  -&  -&  \checkmark& \checkmark &  80.7&77.4\\
         Res-V-CPS$^*$&  -& \checkmark &  -&  -& 81.2&78.3\\
         Seg-V-CPS$^*$&  -& \checkmark &  -&  -&  81.8&78.4\\
         H-CPS$^*$&  \checkmark&  -&  -&  -&  81.4&78.1\\
         H-V-CPS$^*$&  \checkmark&  \checkmark&  -&  -&  82.2&78.9\\
         HDVS&  \checkmark&  \checkmark&  \checkmark&  \checkmark&   \textbf{83.1}&\textbf{79.3}\\
        \hline
    \end{tabular}
    \label{tab:table3}
    \vspace{-5mm}
\end{table*}

\begin{figure*}[t]
    \centering
    \includegraphics[width=1\linewidth]{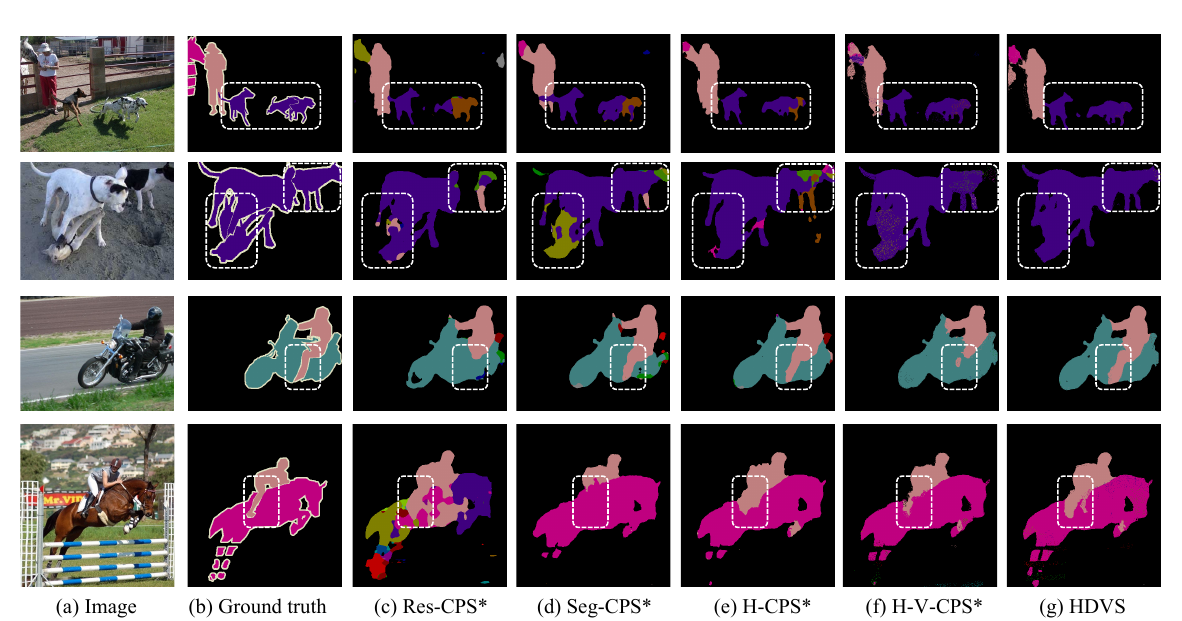}
    \caption{Visualizations of segmentation maps generated by various models, (a) input images to (g) our HDVS model, showcasing progressive enhancements at each step. All models, including the homogeneous baselines Res-CPS$^{*}$/Seg-CPS$^{*}$ (dual convolutional neural network / dual Transformer), utilizing ResNet-50~\cite{he2016deep} as the backbone and employing a combined approach of ResNet-50 and Mit-B2~\cite{xie2021segformer} for heterogeneous structures, are trained using the classic 1464 partition protocol on the PASCAL VOC 2012 train set. H-CPS$^*$ adopts a heterogeneous CNN+Transformer pair, and H-V-CPS$^*$ further introduces a voting module on top of H-CPS$^*$.}
    \vspace{-5mm}
    \label{fig:figure8}
\end{figure*}

\subsection{Ablation Studies}

\textbf{Ablative experiments of different components.} In Table~\ref{tab:table3}, we conduct ablations of HDVS components on PASCAL VOC 2012 under the classic 1464 partition protocol and the blended 1/8 (1323) partition protocol. Specifically, we build on the CPS$^{*}$ dual-branch baseline (a reimplementation of CPS~\cite{chen2021semi} with aligned pseudo-labels filtering, matching hyperparameters, and the same learning-rate schedule) and instantiate: 1) homogeneous baselines Res-CPS$^{*}$/Seg-CPS$^{*}$ (dual CNN / dual Transformer); 2) their ``+FCM only'' variants Res-C-CPS$^{*}$/Seg-C-CPS$^{*}$ (communication branch enabled without voting module) and ``+voting only'' variants Res-V-CPS$^{*}$/Seg-V-CPS$^{*}$; and 3) heterogeneous H-CPS$^{*}$ (CNN+Transformer), H-V-CPS$^{*}$ (voting module with heterogeneous networks), and the full HDVS. All experiments use the ResNet-50 backbone, and Mit-B2 is paired with ResNet-50 in the heterogeneous setting. Table~\ref{tab:table3} shows that 1) on homogeneous baselines, adding FCM yields no gain and can slightly degrade performance relative to the CPS$^{*}$ baselines; 2) adding the voting module consistently improves over the corresponding baselines; and 3) on heterogeneous pairs, H-CPS$^{*}$ outperforms the homogeneous baselines; adding the voting module (H-V-CPS$^{*}$) yields further gains; and the full HDVS (heterogeneous networks + voting + FCM) achieves the best performance. These results show that the voting module and FCM effectively aggregate complementary cues from heterogeneous branches. FCM is effective in our heterogeneous model by transferring complementary global--local cues. However, in common homogeneous dual-branch settings it tends to over-align the branches and reduce diversity.

Figure.~\ref{fig:figure8} visualizes segmentation maps from the representative models described above, trained under the classic 1464 partition protocol, with HDVS showcasing increased robustness in handling complex scenes through diverse components. This is particularly evident when dealing with multiple complex targets, such as the multiple dogs in rows 1 and 2, and in handling areas where objects overlap, such as the legs of characters in rows 3 and 4. The superior performance of H-CPS$^*$ over both Res-CPS$^*$ and Seg-CPS$^*$ demonstrates the synergy between CNNs and Transformers: CNNs capture fine-grained local structures, whereas Transformers model long-range, global semantics, yielding complementary representations. In contrast, as training progresses, homogeneous dual-branch models tend to converge toward similar representations, thereby diminishing the representational diversity critical for effective co-supervision. Moreover, the additional gains of H-V-CPS$^*$ over H-CPS$^*$ indicate that architectural heterogeneity alone is insufficient and pseudo-labels quality remains a limiting factor. By adaptively fusing branch predictions according to confidence, the voting mechanism in H-V-CPS$^*$ suppresses noisy pseudo-labels and improves supervisory consistency, enhancing co-training effectiveness. Finally, HDVS further extends H-V-CPS$^*$ by incorporating a FCM and a dedicated communication branch, enabling implicit feature-level interactions across architectures to transfer complementary knowledge and strengthen collaborative learning.

\textbf{Impact of hyper-parameters.} To evaluate the influence of feature communication intensity on HDVS performance, we conduct comprehensive ablation experiments on the scaling coefficients $\lambda^0$ and $\lambda^1$ in FCM for both the Transformer and CNN branches. We employ ResNet-50 and Mit-B2 as our backbone and train on the classical partition of 1464 images from the PASCAL VOC 2012 train set. As shown in Table~\ref{tab:table6}, by varying one coefficient while keeping the other fixed, we observe that the best performance is achieved with $\lambda^0 = 0.10$ and $\lambda^1 = 0.01$. CNNs and Transformers naturally complement each other: CNNs excel at capturing local details, while Transformers model global dependencies. However, excessive Transformer features disrupt the CNN's representations. Reducing $\lambda^1$ allows the Transformer's global information to integrate gradually, providing guidance without destabilizing the CNN. Consequently, the optimal ratio from Transformer to CNN is set lower than the other.

\begin{table}
\caption{Ablation study on scaling coefficients $\lambda^0$ and $\lambda^1$ using a combination of ResNet-50 and Mit-B2 as the backbone within the classical partition of 1464 training images from the PASCAL VOC 2012 train dataset.}
\label{tab:table6}
\centering
\renewcommand{\arraystretch}{1.4}
\scalebox{0.9}{
\begin{tabular}{>{\centering\arraybackslash}p{1.5cm}>{\centering\arraybackslash}p{1.5cm}>{\centering\arraybackslash}p{3.5cm}}
\hline
{$\lambda^0$}& {$\lambda^1$}& {Performance (\%)} \\
\hline
1.000 & 0.010 & 82.4  \\
0.100 & 0.010 & \textbf{83.1} \\
0.010 & 0.010 & 82.5  \\
0.001 & 0.010 & 82.0  \\
\hline
0.100 & 1.000 & 81.9  \\
0.100 & 0.100 & 82.3 \\
0.100 & 0.010 &  \textbf{83.1}  \\
0.100 & 0.001 & 81.7  \\
\hline
\end{tabular}}
\vspace{-6mm}
\end{table}

\begin{figure}[t]
    \centering
    \scalebox{1}{\includegraphics[width=1\linewidth]{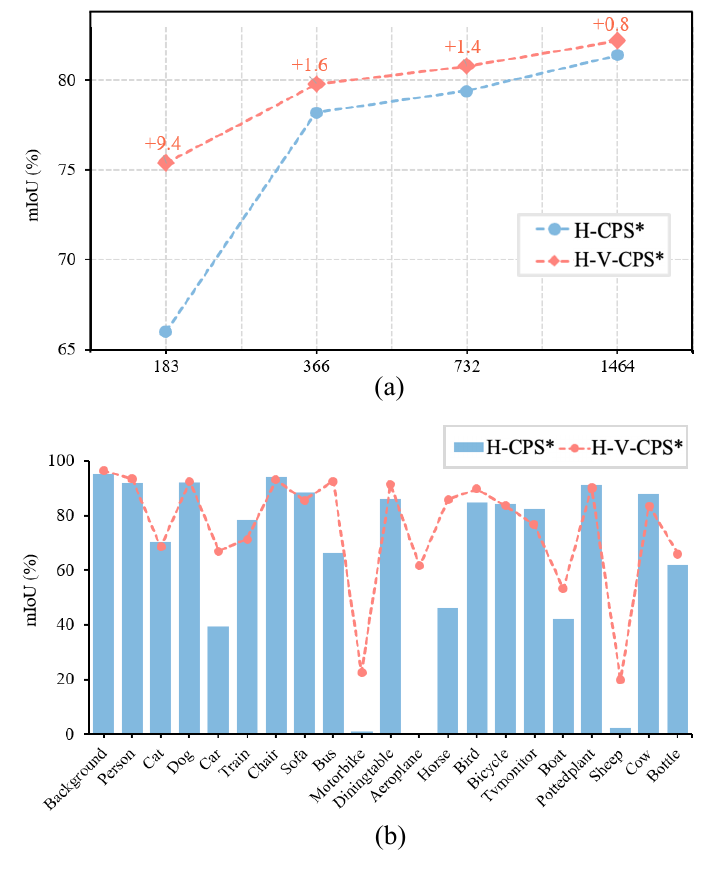}}
    \vspace{-6mm}
    \caption{Effectiveness of the voting module under heterogeneous networks. (a) The performance of H-CPS$^*$ and H-V-CPS$^*$, using a combination of ResNet-50 and Mit-B2 as the backbone, under different numbers of labeled images on the PASCAL VOC 2012 dataset. (b) Under the classical partition protocol with 183 training samples on the PASCAL VOC train dataset, the performance of H-CPS$^*$ and H-V-CPS$^*$ across different categories. mIoU denotes mean intersection over union.}
    \label{fig:figure7}
    \vspace{-2mm}
\end{figure}

\textbf{Impact of the voting mechanism.} To comprehensively evaluate the efficacy of the voting mechanism, we conduct experiments on H-CPS$^*$ and H-V-CPS$^*$ using four different quantities of labeled data under the classical setting: 183, 366, 732, and 1464. Both models use a combination of ResNet-50 and Mit-B2 as their backbone. As illustrated in Fig.~\ref{fig:figure7}(a), we observe a more pronounced impact of the voting module on the model as the labeled data decreases. Notably, there is a 9.4\% increase in accuracy for H-V-CPS$^*$ compared to H-CPS$^*$ in the 183 dataset partition. Fig.~\ref{fig:figure7}(b) illustrates instances where H-CPS$^*$ encounters difficulties in precisely identifying specific individual classes, as evidenced by the near-zero mIoU performance in those categories. This discrepancy stems from the single-model pseudo-label generation's flaws, which may amplify confirmation bias and degrade performance. HDVS uses a voting mechanism for collaborative learning, merging both models' predictions to produce superior pseudo-labels at uncertain pixels, thus effectively solving this problem.

\textbf{Ablations on feature communication mechanisms.} In Table~\ref{tab:fcm_variants}, we compare three communication schemes under identical backbones and training using the classic 1464 and the blended 1/8 (1323) partition protocol on the VOC12 train dataset: (1) Concat+MLP, which concatenates peer features and applies a per-pixel multi-layer perceptron (MLP); (2) Cross-attention, which fuses branches via attention; and (3) FCM, as summarized, it offers the best accuracy--efficiency trade-off: it improves upon the H-CPS$^\ast$ baseline on both partition protocol and surpasses the two heavier alternatives while using the fewest parameters. We attribute the gains to FCM's design, which confines learnable computation to a single projection with parameter-free alignment, thereby avoiding channel inflation and extra decoding. In contrast, concatenation/MLP and cross-attention add heavy projections and multi-layer mappings, increasing parameter counts and overfitting risk.

\textbf{Entropy drop sensitivity.} We assess the effect of discarding high-entropy pixels by sweeping drop ratios of 0\%, 10\%, 20\%, 30\%, and 40\% on PASCAL VOC 2012 using the classical 1464 partition protocol. As shown in Table~\ref{tab:entropy}, removing a moderate fraction is beneficial: mIoU rises from 78.3\% at 0\% to a peak of 83.1\% at 20\%, and remains high at 82.2\% at 30\%. In contrast, both a smaller ratio (10\%, 81.8\%) and a larger ratio (40\%, 79.4\%) reduce HDVS's performance. Overall, the curve saturates near 20\% and is flat over 10$\sim$30\%, indicating robustness and motivating 20\% as a principled default rather than a hand-tuned choice.

\textbf{Parameters and computational overhead.} Table~\ref{tab:params} summarizes the trainable parameters. Overall, HDVS totals 71.0 M parameters (ResNet-50~\cite{he2016deep}: 43.6 M; Mit-B2~\cite{xie2021segformer}: 27.4 M). FCM contributes 2.9 M (already included), while the voting module has no learnable parameters. Under PASCAL VOC~2012 with \(513\times513\) crops, FCM+voting adds 36.9~GFLOPs per image; voting itself is only approximately \( 1.6\times 10^{-3}\)~GFLOPs, which is negligible relative to the backbone and decoder. The overhead is dominated by the \(1{\times}1\) channel projections used for feature alignment.

\begin{table}[t]
\centering
\caption{Cross-branch communication variants. 
We report mIoU on the PASCAL VOC 2012 train dataset under the classic 1464 and the blended 1/8 (1323) partitioning protocols, along with the parameter count of the feature communication components (M).
H-CPS$^\ast$ serves as the baseline without feature communication. Concat + multi-layer perceptron (MLP) concatenates peer features and projects them with an MLP. Cross-attention fuses features via cross-attention.}
\begin{tabular}{lcccc}
\toprule
 & 1464 & 1323 & Params (M) \\
\midrule
H-CPS$^\ast$ &  81.4&  78.1& - \\
Cross-attention & 80.5 & 76.2 &  29.3  \\
Concat + MLP& 82.1 & 78.3 &  7.0\\
Ours (FCM) &  \textbf{83.1} &  \textbf{79.3} & 2.9 \\
\bottomrule
\vspace{-8mm}
\end{tabular}
\label{tab:fcm_variants}
\end{table}

\subsection{Failure Case Analysis} 
\begin{figure}[h]
    \centering
    \scalebox{1}{
    \includegraphics[width=1\linewidth]{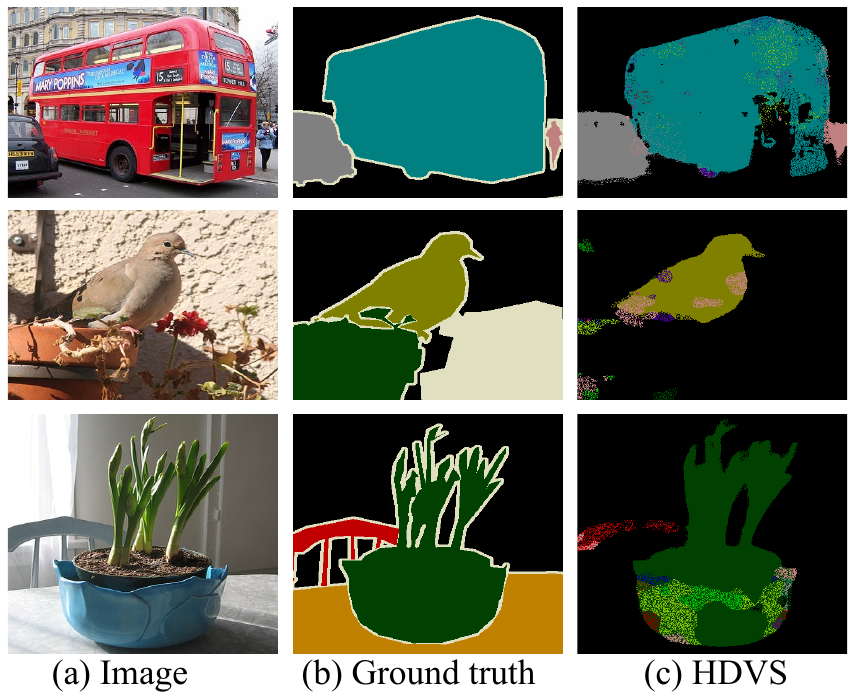}}
    \caption{Failure cases visualizations. From left to right, the categories are (a) input images, (b) ground truth annotations, and (c) segmentation maps generated by our HDVS framework. All models are trained using the 1/16 classical partition protocol on the PASCAL VOC 2012 train dataset.}
    \label{fig:figure9}
    \vspace{-3mm}
\end{figure}

Although HDVS achieves competitive performance on various benchmarks, we observe certain limitations under extreme conditions. These issues are mainly observed when the model is trained with the 1/16 classical partition protocol on the PASCAL VOC 2012 dataset, where the heterogeneous branches may occasionally produce divergent predictions in challenging regions, leading to less satisfactory segmentation. As illustrated in Fig.~\ref{fig:figure9}, examples include the front of a car with drastic color variations (the first row), the subtle boundary between a bird and its background (the second row), and a flowerpot region affected by strong illumination changes (the third row). This behavior stems from noisy pseudo-labels early in training. While the FCM's cross-branch perturbations can be beneficial, under severely limited supervision they may propagate noise across branches and transiently destabilize co-training. Addressing these limitations will be an important direction of our future work.

\begin{table}[t]
\centering
\setlength{\tabcolsep}{5pt}
\caption{HDVS performance under different entropy-drop ratios. All experiments use a combination of ResNet-50 and MiT-B2 as the backbone, within the classical partition of 1464 training images from the PASCAL VOC 2012 train dataset. $\mathrm{\Delta}$ is relative to 20\%.}
\begin{tabular}{lccccc}
\toprule
Drop ratio (\%) & 0 & 10 & 20 & 30 & 40 \\
\midrule
mIoU            & 78.3 & 81.8 & 83.1 & 82.2 & 79.4\\
$\mathrm{\Delta}$ vs.\ 20\% & $-$4.8 & $-$1.3 & - & $-$0.9 & $-$3.7 \\
\bottomrule
\vspace{-5mm}
\end{tabular}
\label{tab:entropy}
\end{table}

\begin{table}[t]
\centering
\caption{Trainable parameter counts (M) of HDVS. The parameter count of the FCM is already included in the student/teacher branch totals and is reported separately for reference only.}
\label{tab:params}
\begin{tabular}{lc}
\hline
Component & Params (M)  \\
\hline
Student branch (ResNet-50)& 43.6   \\
Teacher branch (Mit-B2)& 27.4  \\
FCM  & 2.9 \\
Voting module& 0.0 \\
HDVS & \textbf{71.0} \\

\hline
\end{tabular}

\vspace{-5mm}
\begin{minipage}{0.9\linewidth}
\end{minipage}
\end{table}

\section{Conclusion}
In this paper, we present a novel approach called heterogeneous dual-branch voting supervision (HDVS). HDVS builds upon the cross-supervision architecture, by integrating a heterogeneous dual network structure to enhance the diversity of pseudo-labels, both implicitly and explicitly. Each branch incorporates a feature communication module that enables more effective label extraction at the feature level. Furthermore, a straightforward yet efficient voting mechanism is proposed to refine labels collaboratively generated by the dual branches. This mechanism enables the strengthening of co-confident labels and the weakening of noisy labels. With the collaboration of the two proposed modules, our HDVS achieves significant performance improvement against the state-of-the-art approaches on both the PASCAL VOC 2012 and Cityscapes benchmarks.

\begin{small}
\vspace{.3in} \noindent \textbf{Abbreviations:}
CPS, cross pseudo supervision; EMA, exponential moving average; FCM, feature communication module; H-CPS, heterogeneous cross pseudo supervision; HDVS, heterogeneous dual-branch voting supervision; mIoU, mean intersection over union; Res-CPS, cross pseudo supervision with dual ResNet backbones; Seg-CPS, cross pseudo supervision with dual SegFormer backbones; SOTA, state of the art; SSL, semi-supervised learning; V-CPS, heterogeneous cross pseudo supervision with a voting module.

\vspace{.3in} \noindent \textbf{Data Availability:}
The datasets used in this study are publicly available: the Cityscapes dataset can be downloaded from the official website (\url{https://www.cityscapes-dataset.com/}) and the PASCAL VOC 2012 dataset from the PASCAL VOC project page (\url{http://host.robots.ox.ac.uk/pascal/VOC/voc2012/}); see also the corresponding citations in the paper. Our source code and trained model weights will be made publicly available upon acceptance.

\vspace{.3in} \noindent \textbf{Competing Interests:}
The authors declare that they have no competing interests.

\vspace{.3in} \noindent \textbf{Funding:}
This work was supported by the Fundamental Research Funds for the Central Universities (No. DUT24RC(3)025), and 2024-0011 (No. ZX20240867).

\vspace{.3in} \noindent \textbf{Author contributions:}
Yongqi Shan conceived the method, implemented the framework, and led the experiments and manuscript drafting.
Yunzhi Zhuge contributed to experimental design, analysis, and manuscript revision.
Huchuan Lu supervised the project, advised on methodology, and revised the manuscript.
All authors read and approved the final manuscript.

\begin{acknowledgements}
We would like to express our sincere gratitude to Drs. Pingping Zhang and Lijun Wang for their valuable discussions.
\end{acknowledgements}

\end{small}

\bibliographystyle{naturemag}
\bibliography{reference}

\end{document}